\documentclass[sigconf, screen]{acmart}

\AtBeginDocument{%
  \providecommand\BibTeX{{%
    \normalfont B\kern-0.5em{\scshape i\kern-0.25em b}\kern-0.8em\TeX}}}

\usepackage{multirow}
\usepackage{makecell}
\usepackage{hyperref}
\hypersetup{hidelinks,
	colorlinks=true,
	allcolors=black,
	pdfstartview=Fit,
	breaklinks=true}

\copyrightyear{2022} 
\acmYear{2022} 
\setcopyright{acmcopyright}\acmConference[PIC '22]{Proceedings of the 4th Person in
	Context Workshop}{October 14, 2022}{Lisboa, Portugal}
\acmBooktitle{Proceedings of the 4th Person in Context Workshop (PIC '22), October
	14, 2022, Lisbon, Portugal}
\acmPrice{15.00}
\acmDOI{10.1145/3552455.3555816}
\acmISBN{978-1-4503-9489-5/22/10}
\acmISBN{978-1-4503-9489-5/22/10}

\begin{document}

\title{PIC 4th Challenge: \\
	Semantic-Assisted Multi-Feature Encoding and Multi-Head Decoding for Dense Video Captioning}

\author{Yifan Lu}
\additionalaffiliation{
	\institution{School of Artificial Intelligence, University of Chinese Academy of Sciences}
	\state{Beijing}
	\country{China}
}
\affiliation{%
	\institution{National Laboratory of Pattern Recognition, CASIA}
	\state{Beijing}
	\country{China}
	\postcode{100190}
}
\email{luyifan2021@ia.ac.cn}

\author{Ziqi Zhang}
\affiliation{%
	\institution{National Laboratory of Pattern Recognition, CASIA}
	\state{Beijing}
	\country{China}
	\postcode{100190}
}
\email{zhangziqi2017@ia.ac.cn}

\author{Yuxin Chen} 
%\additionalaffiliation{
%	\institution{School of Artificial Intelligence, University of Chinese Academy of Sciences}
%	\state{Beijing}
%	\country{China}
%}
\authornotemark[1]
\affiliation{%
	\institution{National Laboratory of Pattern Recognition, CASIA}
	\state{Beijing}
	\country{China}
	\postcode{100190}
}
\email{chenyuxin2019@ia.ac.cn}

\author{Chunfeng Yuan}
\authornote{Corresponding author.}
\affiliation{%
	\institution{National Laboratory of Pattern Recognition, CASIA}
	\state{Beijing}
	\country{China}
	\postcode{100190}
}
\email{cfyuan@nlpr.ia.ac.cn}

\author{Bing Li}
\affiliation{%
	\institution{National Laboratory of Pattern Recognition, CASIA}
	\state{Beijing}
	\country{China}
	\postcode{100190}
}
\email{bli@nlpr.ia.ac.cn}

\author{Weiming Hu}
%\additionalaffiliation{
%	\institution{School of Artificial Intelligence, University of Chinese Academy of Sciences}
%	\state{Beijing}
%	\country{China}
%}
\authornotemark[1]
\additionalaffiliation{
	\institution{CAS Center for Excellence in Brain Science and Intelligence Technology}
	\state{Beijing}
	\country{China}
}
\affiliation{%
	\institution{National Laboratory of Pattern Recognition, CASIA}
	\state{Beijing}
	\country{China}
	\postcode{100190}
}
\email{wmhu@nlpr.ia.ac.cn}


\renewcommand{\shortauthors}{Lu, et al.}

\begin{abstract}
The task of Dense Video Captioning (DVC) aims to generate captions with timestamps for multiple events in one video. Semantic information plays an important role for both localization and description of DVC. 
We present a semantic-assisted dense video captioning model based on an encoding-decoding framework. In the encoding stage, we design a concept detector to extract semantic information, which is then fused with multi-modal visual features to sufficiently represent the input video. In the decoding stage, we design a classification head, paralleled with the localization and captioning heads, to provide semantic supervision.
Our method achieves significant improvements on the YouMakeup dataset \cite{wang2019youmakeup} under DVC evaluation metrics and achieves high performance in the Makeup Dense Video Captioning (MDVC) task of \href{http://picdataset.com/challenge/task/mdvc/}{PIC 4th Challenge}. 

\end{abstract}




%  \includegraphics[width=\textwidth]{fig1}
%  \label{fig1}

%\settopmatter{printacmref=false}
\maketitle

\section{Introduction}

Video Captioning (VC) is an important research branch of video understanding. The task of VC aims to generate a natural sentence to describe the content of a video. The VC task only deals with ideal situations where the provided video is short and the generated sentence only describes one main event in the video. However, for most natural videos composed of multiple events, a single sentence cannot cover the content of the video.

% DVC
To tackle this issue, the task of Dense Video Caption (DVC) is developed for temporal localization and description generation for multiple events in one video. Intuitively, DVC can be divided into two sub-tasks, event localization and event captioning. The localization sub-task aims to predict the timestamps of each event. This requires the DVC model to decide temporal boundaries between event and non-event segments, and discriminate one event from another. For the captioning sub-task, the model needs to generate a natural sentence to describe each corresponding event. 

\begin{figure}[t]
 	\centering
 	\includegraphics[width=\linewidth]{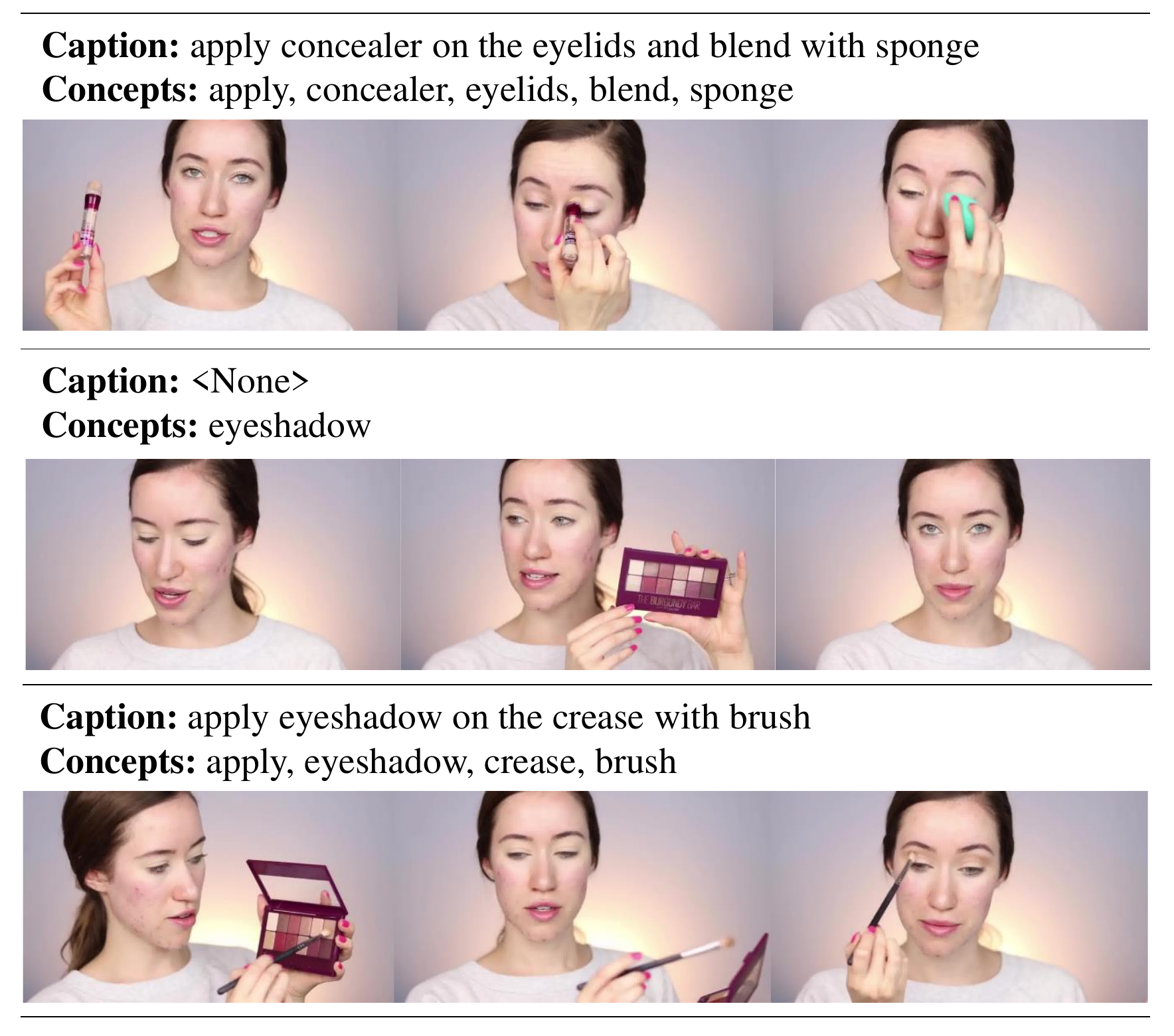}
 	\caption{In the YouMakeup dataset \cite{wang2019youmakeup}, different segments have different concepts. Segments with events (top and bottom rows) have concepts including makeup actions, products, tools, and face areas being affected by makeup actions. Non-event segments may not have concepts of make-up actions and affected face areas.}
 	\Description{Different segments have different concepts.}
 	\label{fig1}
 \end{figure}

Recent works \cite{wang2021end, deng2021sketch} have proposed models that can achieve good performance under DVC metrics. However, semantic information, which is proved to be useful in VC tasks \cite{gan2017semantic, perez2021attentive}, hasn't been used in DVC tasks yet.
% motivation
%Introducing high-level semantic concepts (i.e. object and action tags) of video is proved to be useful in VC tasks \cite{gan2017semantic, perez2021attentive}, for it helps to bridge the semantic gap between video and text. In DVC tasks, 
As shown in Figure \ref{fig1}, we notice that there are different concepts (i.e. actions and object tags) in different segments in one video. This can help the DVC model decide temporal boundaries between different segments. Introducing high-level semantic concepts also helps to bridge the semantic gap between video and text. 
% our model

To make full use of semantic information, we introduce \textbf{semantic assistance} to our model, both in the encoding and decoding stage.   
We use PDVC, which stands for \emph{end-to-end dense \textbf{V}ideo \textbf{C}aptioning with \textbf{P}arallel \textbf{D}ecoding} \cite{wang2021end}, as our baseline model. PDVC is a transformer-based framework with parallel sub-tasks.
In the encoding stage, a \textbf{concept detector} is designed to extract frame-level semantic information. We design a \textbf{fusion module} to integrate all the features. In the decoding stage, a \textbf{classification sub-task} is added in parallel with localization and captioning sub-tasks. By predicting attributes for events, the classification sub-task can provide event-level semantic supervision. 
Experimental results show that our strategy of using semantic information achieves significant improvement on the YouMakeup dataset \cite{wang2019youmakeup} under DVC evaluation metrics.
  
\section{Related Works}
DVC models often follow the encoder-decoder framework. The encoder extracts visual features from the raw video and gives a general representation of the video. Off-the-shelf models, such as C3D \cite{ji20123d}, I3D \cite{carreira2017quo}, and ResNet \cite{he2016deep} can be used as the backbones of the encoder. The decoder takes the encoding visual representation as input and performs two tasks including event localization and event captioning. 
%The relationship between 2 sub-tasks is varied. 

Krishna et al. \cite{krishna2017dense} propose the first DVC captioning model with a two-stage framework. The decoder combines a proposal module and a captioning module. The proposal module performs the localization sub-task by selecting numerous video segments as event proposals, then the captioning module generates captions for each proposal. Motivated by transformer-based end-to-end object detection methods \cite{carion2020end, zhu2020deformable}, Wang et al. \cite{wang2021end} propose a parallel decoding method where the DVC task is considered as a set prediction problem. An event set with temporal locations and captions is directly predicted by applying localization and captioning sub-task in parallel. Deng et al. \cite{deng2021sketch}, in another way, reverse the "localize-then-captioning" fashion and propose a top-down scheme. In their method, a paragraph is firstly generated to describe the input video from a global view. Each sentence of the paragraph is treated as an event and then temporally grounded to a video segment for fine-grained refinement. 

\begin{figure*}[t]
  \includegraphics[width=\textwidth]{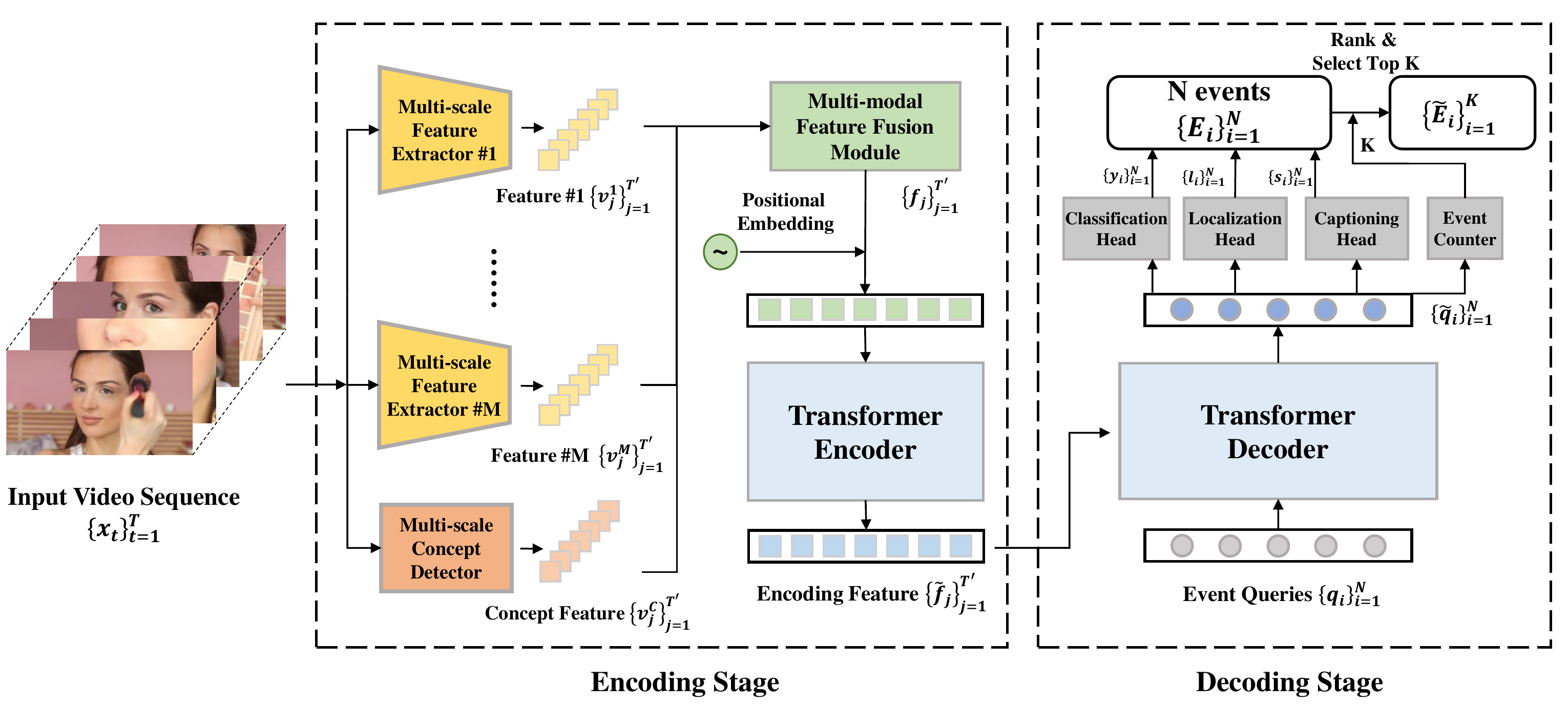}
  \caption{Overview of our proposed DVC model. M multi-scale feature extractors and a multi-scale concept detector are respectively used to extract frame-level multi-modal visual features and the concept feature from video frame sequences, which are then fused by the multi-modal feature fusion module. The transformer encoder is used to obtain the final representation of the video. The transformer decoder and four parallel heads are proposed to predict the labels, locations, captions, and the number of events.}
  \Description{Overview of the proposed DVC model}
  \label{fig2}
\end{figure*}

\section{Methods}
\subsection{Overview}
In the DVC task, given an input video sequence $\{ \boldsymbol{x}_t \}^T_{t=1}$, the model needs to predict all the $K$ events $\{ \boldsymbol{\tilde{E}}_i|\boldsymbol{\tilde{E}}_i=(\boldsymbol{l}_i, \boldsymbol{s}_i) \}^K_{i=1}$. $\boldsymbol{l}_i$ and $\boldsymbol{s}_i$ respectively stand for timestamps and caption sentences of the \emph{i}-th event.
In our work, PDVC \cite{wang2021end} is used as the baseline model. We further add a semantic concept detector, a multi-modal feature fusion module and a classification head on the basis of PDVC. Here we present an overview of our model. 

% encoding
As shown in Figure \ref{fig2}, our model follows the encoder-decoder pipeline. In the encoding stage, a video frame sequence is fed into $M$ multi-scale feature extractors and a multi-scale concept detector. The multi-modal feature fusion module is employed to fuse all the extracted features. The transformer encoder takes the fused feature sequence with positional embedding to produce the final visual representation.
% decoding

In the decoding stage, the transformer decoder takes event query sequences and encoding feature as input, followed by four parallel heads. The localization and captioning heads predict the timestamps and captions for each query respectively. The classification head performs a multi-label classification task to assign each event to predefined classes. The event counter predicts the actual number of events in the video.

\subsection{Feature Encoding}
\subsubsection{Multi-scale feature extractor}
$M$ multi-scale feature extractors take video frame sequence $\{\boldsymbol {x}_t \}^T_{t=1}$ to extract features of $M$ modalities $\{ \boldsymbol{v}_j^m \}^{T'}_{j=1}$, where $m = 1,...,M$.
Each multi-scale feature extractor is composed of an off-the-shelf pretrained feature extractor (e.g. Swin Transformer \cite{liu2021swin}, I3D \cite{carreira2017quo}) and a following temporal convolution network with $L$ layers. Multi-scale features are obtained by temporally concatenating the raw features with outputs of $L$ 1D temporal convolution layers (stride=2). Thus the output sequence length $T'$ can be calculated as:
\begin{equation}
	T' = \sum_{l=0}^{L} \lceil \frac{T}{2^l} \rceil .
\end{equation}

\subsubsection{Concept Detector}
 The concept detector is a pretrained module to predict concept vectors $\{ \boldsymbol{c}_t \}^{T}_{t=1}$, i.e. the probabilities of concepts appearing in each video frame. The concept detection approach is defined as follows. We first use NLTK toolkit \cite{bird2009natural} to apply part-of-speech tagging to each word in the training corpus. We choose nouns and verbs of high word frequency as $N_c$ concepts. For the \emph{t}-th frame with captions, its ground truth concept vector $\boldsymbol{c}_t = [c^1_t, ..., c^{N_c}_t] $ is assigned by:
\begin{equation}
    c^i_t = \left\{ \begin{array}{ll}
	1 & \text{if \emph{i}-th concept in the caption}\\
%	\\ & \text{corresponding caption of frame t} \\
	0 & \text{otherwise}\\
\end{array} \right.
, i = 1,2,...N_c .
\end{equation}
The concept detector contains a pretrained feature extractor and a trainable multi-layer perceptron. Frames without captions (i.e. non-event frames) are not taken into consideration at training stage. 

In the whole DVC pipeline, the pretrained concept detector serves as a feature extractor for frames both with and without captions. A temporal convolution network also follows to produce multi-scale feature $\{ \boldsymbol{v}^C_j \}^{T'}_{j=1}$ from concept vectors $\{ \boldsymbol{c}_t \}^{T}_{t=1}$.

\subsubsection{Multi-Modal Feature Fusion Module} 
The multi-modal feature fusion module fuses features from all modalities, as well as the concept feature. Features are projected into embedding space and then concatenated by frame. The fused feature is denoted as $\{\boldsymbol{f}_j\}^{T'}_{j=1}$.

\subsubsection{Transformer Encoder} 
The transformer encoder takes the fused feature sequence $\{\boldsymbol{f}_j\}^{T'}_{j=1}$ with positional embedding to produce final visual representation $\{\boldsymbol{\tilde{f}}_j\}^{T'}_{j=1}$ by applying multi-scale deformable attention (MSDatt) \cite{zhu2020deformable}. MSDatt helps to capture multi-scale inter-frame interactions.

\subsection{Parallel Decoding}
The decoding part of the model contains a transformer decoder and four parallel heads. The transformer decoder takes $N$ event queries $\{\boldsymbol{q}_i\}^{N}_{i=1}$ and encoding frame-level feature $\{\boldsymbol{\tilde{f}}_j\}^{T'}_{j=1}$. Each event query corresponds with a video segment. The transformer decoder also applies MSDatt to capture frame-event and inter-event interactions. Four heads make predictions based on the output event-level representations $\{\boldsymbol{\tilde{q}}_i\}^{N}_{i=1}$ of transformer decoder.

\subsubsection{Localization head} 
The localization head predicts the timestamps $\{\boldsymbol{l}_i\}^N_{i=1}$ of each query using a multi-layer perceptron. Each timestamp contains the normalized starting and ending times. 

\subsubsection{Captioning head}
 The captioning head employs a LSTM network to predict caption sentences $\{\boldsymbol{s}_i\}^{N}_{i=1}$ of each query. For \emph{i}-th query, the event level representation $\boldsymbol{\tilde{q}}_i$ is fed into LSTM every time step and a fully-connected layer takes the hidden state of LSTM to predict words.

\subsubsection{Classification head}
 Each ground truth event is assigned with labels that indicate certain attributes of the event. The classification head predicts the label vector $\{\boldsymbol{y}_i\}^{N}_{i=1}$. The head is composed of a multi-layer perceptron. Each value of vector $\boldsymbol{y}_i$ indicates the probability of a certain label in the event. The classification sub-task, which brings semantic supervision to the model, serves as an auxiliary task for DVC.

\subsubsection{Event counter} The event counter predicts the actual number of events in the video by performing a multi-class classification. The counter contains a max-pooling layer and a fully-connected layer, taking $\boldsymbol{\tilde{q}}_i$ and predicting a vector $k_{num}$ of probabilities of the certain numbers. The length of $k_{num}$ is set to be the expected max number of events plus 1. The actual event number is obtained by $K = \text{argmax} (k_{num})$.

\subsection{Training and Inference}
\subsubsection{Training} \ 
% pretrained
In the training stage, we fix the parameters of the pretrained feature extractors and the concept detector. Feature extractors are directly loaded with off-the-shelf pretrained parameters. The concept detector is offline trained using focal loss \cite{lin2017focal}  to alleviate the problems of unbalanced samples.

% matching
When training the whole DVC model, the predicted event set $\{E_i\}^N_{i=1}$ has to be matched with the ground truths. We use the Hungarian algorithm to find the best matching, following \cite{wang2021end}. 

% loss
The captioning loss $L_c$ and localization loss $L_l$ are calculated only using matched queries. $L_c$ is the cross-entropy between the ground truth and the predicted probabilities of words. $L_l$ is the gIOU loss \cite{rezatofighi2019generalized} between matched prediction and ground truth pairs. The classification loss $L_{cls}$ is calculated using focal loss between all predicted labels and their targets. For the matched queries, the label target is equal to the matched ground truth. For the unmatched queries, the label target is set to be an all-zero vector. The counter loss is the cross-entropy between the predicted result and the ground truth. The DVC loss is the weighted sum of the four losses above. 

\subsubsection{Inference}
% rank and select 
In the inference stage, the predicted $N$ event proposals $\{E_i\}^N_{i=1}$ are ranked by confidence. Following \cite{wang2021end}, the confidence is the sum of the classification confidence and the captioning confidence. The top $K$ events are chosen as the final DVC result $\{ \tilde{E_i}\}^K_{i=1}$.

\section{Experiments}
\subsection{Settings}
\subsubsection{Dataset} 
We conduct experiments on the YouMakeup dataset \cite{wang2019youmakeup}. The YouMakeup dataset contains 2800 make-up instructional videos of which the length varies from 15s to 1h. There are a total of 30,626 events with 10.9 events on average for each video. Each event is annotated with a caption, a timestamp, and grounded facial area labels from 25 classes. We follow the official split with 1680 for training, 280 for validation, and 840 for test.

\subsubsection{Evaluation Metrics} \
We evaluate our method using the evaluation tool provided by
the 2018 ActivityNet Captions Challenge in aspects of localization and caption. 
For localization performance, we compute the average precision (P) and recall (R) across tIoU thresholds of 0.3/0.5/0.7/0.9. For captioning performance, we calculate BLEU4 (B4), METEOR (M), and CIDEr (C) of the matched pairs between generated captions and the ground truth across tIOU thresholds of 0.3/0.5/0.7/0.9.

\subsubsection{Implementation details} \
We use PDVC \cite{wang2021end} as our baseline model.
Pretrained I3D \cite{carreira2017quo} and Swin Transformer (Base) \cite{liu2021swin} are used to extract frame-level motion and appearance features. The concept detection is performed on Swin Transformer feature of every frame, and the concept number $N_c$ is set to 100. For parallel computing, all the feature sequences are temporally resized into the same length. Sequences with a length larger than 1024 are temporally interpolated into the length of 1024. Those of length less than 1024 are padded to 1024. In the decoding stage, the grounded facial area labels are predicted by the classification head. The number of queries $N$ and the length of $k_{num}$ are set to 35 and 11. Other settings follow the baseline model PDVC. 

\begin{table}%[htbp]
  \begin{tabular}{ccccccc}
	    \toprule
	    Method & Dataset &P & R & B4 & M & C  \\
	    \midrule
		PDVC \cite{wang2021end} & \multirow{2}{*}{val} & 31.47 & 23.76 & 6.30 & 12.49 & 68.18 \\
		ours & &  \textbf{48.80 } & \textbf{29.28} & \textbf{14.24} & \textbf{22.01} & \textbf{137.23} \\
		\hline
		PDVC \cite{wang2021end} & \multirow{2}{*}{test} & 32.23 & 24.82 & 5.72 & 12.25 & 65.54 \\
		ours &  & \textbf{48.20} & \textbf{28.04} & \textbf{13.91} & \textbf{21.56} & \textbf{135.45} \\
	  \bottomrule
	\end{tabular}
	 \caption{Evaluation Results on validation and test dataset compared with baseline}
	 \label{tab:base}
	\end{table}

\begin{table}
	\begin{tabular}{ccccccc}
	\toprule
	Feature & Fusion &P & R & B4 & M & C  \\	
	\midrule
	i3d & - & 43.05 & 30.05 & 10.20 & 17.89 & 111.23 \\
	swin & - & 43.34 & 29.51 & 10.98 & 15.78 & 108.76 \\
	i3d+swin & early & 47.70 & \textbf{32.40} & 13.25 & 20.25 & 122.97\\
	i3d+swin & late & \textbf{48.26} & 32.12 & \textbf{13.75} & \textbf{20.55} & \textbf{130.14} \\
	\bottomrule
	\end{tabular}
	\caption{Ablation study: feature fusion}
	\label{tab:features}
\end{table}
\begin{table}%[htbp]
	\begin{tabular}{ccccccc}
	\toprule
	\makecell[c]{Concept \\ detector} &
	\makecell[c]{Classification \\ head} &
	\makebox[0.015\textwidth][c]{P} &
	\makebox[0.015\textwidth][c]{R} &
	\makebox[0.015\textwidth][c]{B4} &
	\makebox[0.015\textwidth][c]{M} &
	\makebox[0.015\textwidth][c]{C} \\
	\midrule
	- & - & 48.26 & 32.12 & 13.75 & 20.55 & 130.14 \\
	\checkmark & - & 47.71 & \textbf{32.62} & 14.10 & 21.64 & 132.50 \\
	- & \checkmark & 45.13 & 27.08 & 13.75 & 21.01 & 128.51  \\
	\checkmark & \checkmark & \textbf{48.80} & 29.28 & \textbf{14.24} & \textbf{22.01} & \textbf{137.23}\\
	\bottomrule
	\end{tabular}
	\caption{Ablation study: semantic assistant}
	\label{tab:semantic}
\end{table}
\begin{table}%[htbp]
	\begin{tabular}{ccccccc}
	\toprule
	 \makecell[c]{Max event \\ number} &
	 \makebox[0.01\textwidth][c]{Data split} &
	 \makebox[0.01\textwidth][c]{P} &
	 \makebox[0.01\textwidth][c]{R} &
	 \makebox[0.01\textwidth][c]{B4} &
	 \makebox[0.01\textwidth][c]{M} &
	 \makebox[0.01\textwidth][c]{C} \\
	\midrule
	10 & all & 48.80 & \textbf{29.28} & 14.24 & 22.01 & 137.23 \\
	7 & all & 48.50 & 23.57 & 14.18 & 22.71 & 144.80 \\
	5 & all & \textbf{49.51} & 20.50  & \textbf{14.85} & 23.69 & \textbf{157.57} \\
	3 & all & 48.16 & 14.50 & 13.47 & \textbf{24.21} & 151.67 \\
	\hline
	3 & num>3 & 51.24 & 12.28 & 13.64 & 24.77 & 165.76 \\
	3 & num<=3 & 40.12 & 13.30  & 12.41 & 20.65 & 61.92 \\
	\bottomrule
	\end{tabular}
	\caption{Ablation study: different max event number}
		\label{tab:counter}
\end{table}

\subsection{Comparison with baseline}
Table \ref{tab:base} shows DVC metrics on validation and test dataset. Our methods achieves  55.07\%/23.23\%/126.03\%/76.22\%/101.28\% relative gains on validation dataset and 49.55\%/12.97\%/143.18\%/76.00\%/106.67\% on test dataset under the metrics of P/R/B4/M/C compared with the baseline model respectively.

\subsection{Ablation study}
% feature fusion
\subsubsection{Feature fusion}
We evaluate the effectiveness of the usage of multi-modal features on the validation set. We also try early feature fusion. Instead of fusing multi-scale features, features are fused before the temporal convolution network. As shown in Table \ref{tab:features}, using multi-modal features helps to improve all the 5 DVC metrics in comparison with only using feature of one modality. Compared with early fusion, the late fusion method has higher precision and captioning scores but slightly lower recall. The results demonstrate that: 1) Using multi-modal features helps to improve model performance. 2) Details can be better captured by applying late fusion on multi-scale features.

% semantic assistant
\subsubsection{Semantic assistance} 
We evaluate the effectiveness of two semantic assistance modules on the validation set. Table \ref{tab:semantic} shows that: 1) Adding the concept detector increases recall and captioning scores; 2) The classification sub-task cannot bring performance gain alone; 3) Better precision and caption scores can be obtained by applying the concept detector and classification head together.

% event counter
\subsubsection{Expected max event number}
 We try different settings of the expected max event number, which is the upper bound of the event counter output $K$. Table \ref{tab:counter} shows that as the max event number decreases, precision
and captioning scores increase but recall decreases. We also split the validation into 2 parts by event number. When setting the max event number to 3, the model has higher precision and captioning scores on videos containing more than 3 events, oppositely on videos with no more than 3 events. Results can be explained by the trade-off between precision and recall. Since BLEU4/METEOR/CIDEr are only computed on events tIOU-matched with the ground truths, captioning scores are positively correlated with the precision score.
\section{Conclusion}
In this paper, we present a semantic-assisted dense video captioning model with multi-modal feature fusion. The concept detector extracts semantic feature that is fused with other multi-modal visual features. The classification sub-task provides semantic supervision. Experiments prove that our method achieves significant performance on DVC tasks. 

\begin{acks}
This work was supported by Beijing Natural Science Foundation (Grant No. JQ21017, 4224091), the Natural Science Foundation of China (Grant No. 61972397, 62036011, 62192782, 61721004), the Key Research Program of Frontier Sciences, CAS (Grant No. QYZDJ-SSW-JSC040), and the China Postdoctoral Science Foundation (Grant No. 2021M693402).
\end{acks}

%\begin{table}
%  \caption{Frequency of Special Characters}
%  \label{tab:freq}
%  \begin{tabular}{ccl}
%    \toprule
%    Non-English or Math&Frequency&Comments\\
%    \midrule
%    \O & 1 in 1,000& For Swedish names\\
%    $\pi$ & 1 in 5& Common in math\\
%    \$ & 4 in 5 & Used in business\\
%    $\Psi^2_1$ & 1 in 40,000& Unexplained usage\\
%  \bottomrule
%\end{tabular}
%\end{table}
%
%
%
%\begin{table*}
%  \caption{Some Typical Commands}
%  \label{tab:commands}
%  \begin{tabular}{ccl}
%    \toprule
%    Command &A Number & Comments\\
%    \midrule
%    \texttt{{\char'134}author} & 100& Author \\
%    \texttt{{\char'134}table}& 300 & For tables\\
%    \texttt{{\char'134}table*}& 400& For wider tables\\
%    \bottomrule
%  \end{tabular}
%\end{table*}
%
%\begin{math}
%  \lim_{n\rightarrow \infty}x=0
%\end{math},
%
%\begin{equation}
%  \lim_{n\rightarrow \infty}x=0
%\end{equation}
%
%\begin{displaymath}
%  \sum_{i=0}^{\infty} x + 1
%\end{displaymath}
%and follow it with another numbered equation:
%\begin{equation}
%  \sum_{i=0}^{\infty}x_i=\int_{0}^{\pi+2} f
%\end{equation}
%
%\section{Figures}
%
%\begin{figure}[h]
%  \centering
%  \includegraphics[width=\linewidth]{sample-franklin}
%  \caption{1907 Franklin Model D roadster. Photograph by Harris \&
%    Ewing, Inc. [Public domain], via Wikimedia
%    Commons. (\url{https://goo.gl/VLCRBB}).}
%  \Description{A woman and a girl in white dresses sit in an open car.}
%\end{figure}

\bibliographystyle{ACM-Reference-Format}
\bibliography{sample-base}

\end{document}